\documentclass{article}

\usepackage{arxiv}

\usepackage[utf8]{inputenc} 
\usepackage[T1]{fontenc}    
\usepackage[colorlinks]{hyperref}       
\usepackage{url}            
\usepackage{booktabs}       
\usepackage{amsfonts}       
\usepackage{nicefrac}       
\usepackage{microtype}      
\usepackage{lipsum}
\usepackage{graphicx}
\usepackage{amsmath}

\makeatletter

\let\c@table\c@figure
\makeatother 

\title{Combining expert knowledge and neural networks to model environmental stresses in agriculture}

\author{
  Kostadin Cvejoski \\
  Competence Center Machine Learning Rhine-Ruhr \\
  53757 Sankt Augustin, Germany \\
  \texttt{kostadin.cvejoski@iais.fraunhofer.de} 
   \And
  Jannis Schuecker \\
  Fraunhofer IAIS \\
  53757 Sankt Augustin, Germany \\
  \texttt{jannis.schuecker@iais.fraunhofer.de} 
  \And
  Anne-Katrin Mahlein \\
  Institute for Sugar Beet Research \\
  Goettingen, Germany \\
  \texttt{Mahlein@ifz-goettingen.de} 
  \And
  Bogdan Georgiev \\
  Competence Center Machine Learning Rhine-Ruhr and \\ Fraunhofer IAIS
  53757 Sankt Augustin, Germany \\
  \texttt{bogdan.georgiev@iais.fraunhofer.de} 
}

\begin{document}
\maketitle

\begin{abstract}
    In this work we combine representation learning capabilities of neural network with agricultural knowledge from experts to model environmental heat and drought stresses. We first design deterministic expert models which serve as a benchmark and inform the design of flexible neural-network architectures. Finally, a sensitivity analysis of the latter allows a clustering of hybrids into susceptible and resistant ones. 
\end{abstract}

\keywords{neural networks \and deterministic expert-driven models \and  informed machine learning \and agriculture \and representation learning}

\section{Introduction}

The population of the earth is constantly growing and therefore also the demand for food. In consequence, breeding crop plants which most efficiently make use of the available cropland is one of the greatest challenges nowadays. In particular, plants which are resilient and resistant to environmental stresses are desirable. The development of such plants relies on the investigation of the interaction between the plant's genes and the environmental stresses. In order to be able to investigate the interaction a quantitative representation of the environmental stresses is needed. Here, we consider this representation combining state-of-the-art data-driven methods with expert-driven modeling from agriculture.

Briefly put, it has been reported that environmental stress such as inappropriate or extreme temperature conditions, lack of sufficient moisture, etc., can significantly impede the life cycle development of corn, thus leading to yield reductions (cf. \cite{AEB, NN, L} and the references therein). Moreover, there appears to be a period of fundamental importance (approx. around the plant's pollination) at which negative environmental impacts can be critical. Namely, it is speculated that heat and drought (or a combination of the two, abiotic stress) represent significant yield reduction factors provided they appear at vital time frames.  

Modelling the effects of heat and drought stresses on corn yields is a well-studied problem - ranging from straight-forward models based on a few (climate-dependent) environmental features to complex predictions encompassing highly non-linear effects and quantities such as photosynthesis, grain filling rates, daily biomass productions, etc. For a thorough and accessible survey we refer to \cite{Zhe1}.

Recent progress in artificial intelligence has been largely driven by deep neural networks learning various tasks ranging from image classification \cite{INET}, medical diagnosis \cite{MED}, to game AI \cite{GO}. The internal layers of neural networks implicitly learn a representation of the data which is suitable for the current task. Therefore, using neural networks to learn a representation of environmental stresses is an appealing approach. A purely end-to-end approach using neural network, however, constitutes a 'black box' with the lack of interpretability.   

In this work we will leverage the advantages of both worlds, i.e., the expert knowledge gathered by decades of research as well as the flexibility and representation learning abilities of neural networks. In order to combine the two worlds we follow an informed machine learning approach, i.e., we incorporate expert knowledge in the machine learning model through feature-engineering and -selection and model selection. For extensive overview over the informed machine learning approaches we refer to \cite{vonrueden2020informed}.

\section{Methodology and Theory}

In this section we describe the methods we use to solve the two objectives, i.e., the stress design and the regression on the yield reduction as well as the clustering of susceptible/resistant hybrids.

\subsection{Basic architecture}

\begin{figure*}[!ht]
\centering 
\includegraphics[width=6cm]{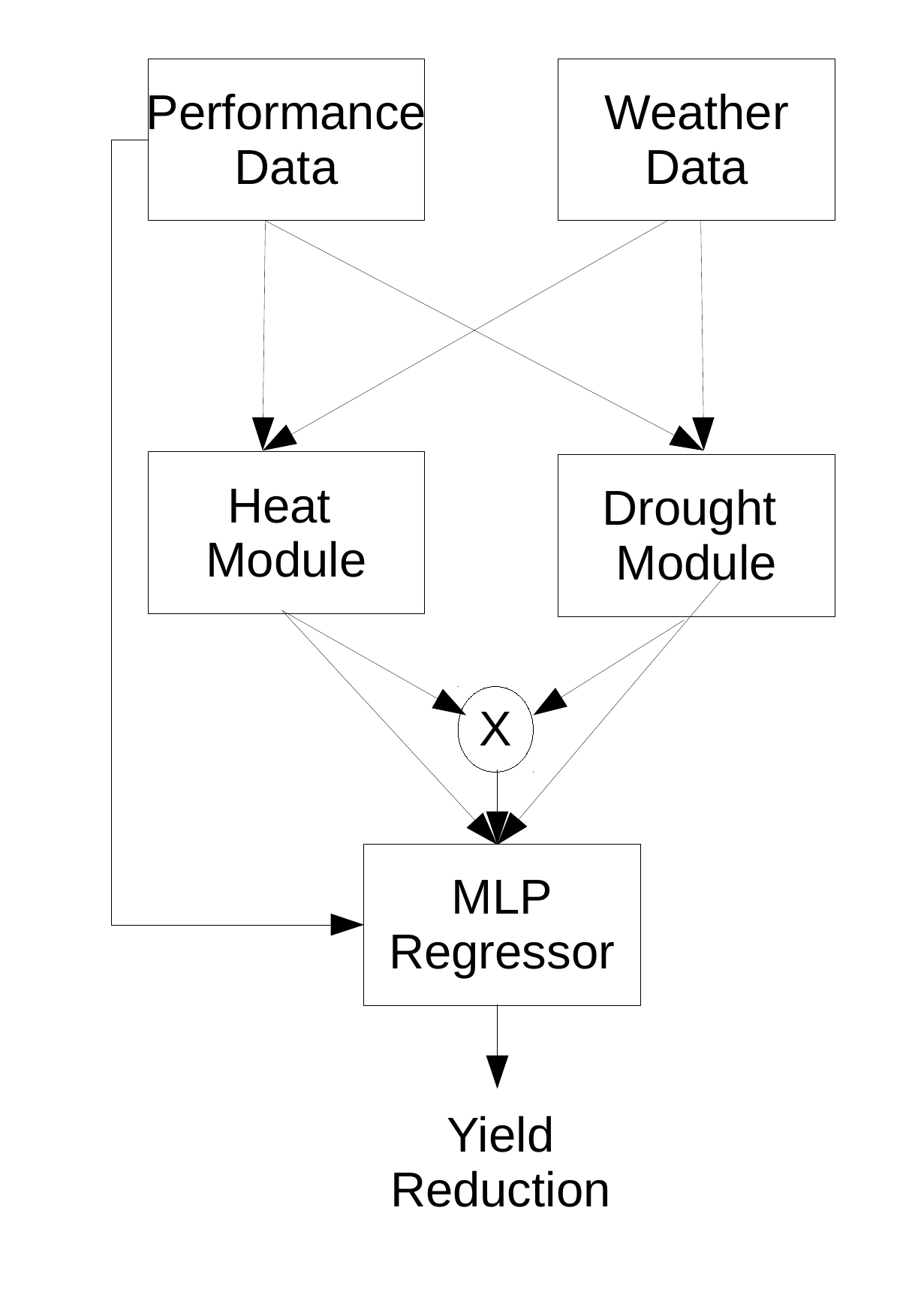}
\caption{Basic architecture of our models}\label{fig:arch}
\end{figure*}

Figure \ref{fig:arch} shows the basis architecture of our approach. The performance data and the weather data serve as inputs to the Heat (H) and Drought (D) modules, where the stress representations are computed. Here we choose between two different kinds of models, deterministic expert models (DEM) and convolutional neural netowks (CNN). Subsequently, the stress representations serve as input to a multilayer perceptron to predict the yield reduction. We refer to the complete models as DEM-MLP and CNN-MLP, respectively. The individual parts will be presented in detail in the following.

\subsection{Stress representations}
\subsubsection{Deterministic stress models}\label{DEM-models}

Using the well-known APSIM platform as a starting point (cf. \cite{Zhe1} and the references therein), we start by proposing a couple of deterministic stress models driven mainly by expert knowledge and agricultural results. In subsequent sections we refer to this models as DEM (deterministic expert models).

On one hand, we roughly model heat stress by means of appropriate cut-off functions applied to the daily temperature values (cf. Subsection \ref{subsec:appendix-det-heat} for further details).

On the other hand, our initial drought stress model is more involved and captures most of the environment/performance data at hand, e.g. utilizing features such as precipitation, soil dynamics, irrigation, humidity, elevation, etc. A central concept here is the model of evapotranspiration (ET) based on a Penman-Monteith type of equation (cf. \cite{APR}). For further details we refer to Subsection \ref{subsec:appendix-det-drought}.

\subsubsubsection{\textbf{Growing Degree Units (GDU), Growth Stages and Time Period Partitions}} \label{subsec:GDUs}

A substantial part of our deterministic analysis is based on the concept of growing degree units (GDU). We observe that the planting instances occur over different time frames (i.e. planting/harvest days may differ for different planting instances) and, furthermore, corn plants could mature at different rates. However, we assume that each corn plant undergoes similar well-established growth/maturity stages such as emergence, rapid growth, pollination, grain fill, etc, with additional sub-stages such as tasseling and silking (cf. \cite{NN, AEB}). 

Having the above motivation in mind, we utilize the notion of growing degree units (GDU) - a tool to measure corn growth stages. The daily GDU quantity is computed as:

\begin{equation} \label{eq:daily-GDU}
    GDU := \frac{T_{max} - T_{min}}{2} - T_{base},
\end{equation}

where $T_{max}, T_{min}$ correspond to the daily maximum and minimum temperatures and where $T_{base}$ is a certain threshold set to $10 \, C^o$ (in the present case for corn plants).

Further, one assumes that the corn has entered a certain growth stage, provided the accumulated GDU has reached an appropriate corresponding threshold - e.g. according to \cite{NN}, the pollination period is reached at around $1135$ accumulated GDUs.

Of course, one should keep in mind that different corn hybrid could mature with different rates - this should be reflected in the final growth stage analysis.

Finally, we have used the sources (cf. \cite{AEB, RWG, L, ET, WHN} and references therein) to identify and sub-refine the needed GDU accumulation thresholds for each growth stage. For further motivation and discussion on our GDU accumulation refinement, we refer to Subsection \ref{subsec:appendix-GDUs}.

\subsubsubsection{\textbf{Deterministic heat stress}}

A straight-forward daily heat stress model we utilize is:

\begin{equation} \label{eq:det-heat-stress}
    S_H := 1 - \begin{cases} 0, \quad \text{if } T_{mean} \geq T_3,  \\
    1, \quad T_{mean} \leq T_2, \\
    \frac{T_3 - T_{mean}}{T_3 - T_2}, \quad \text{if } T_2 < T_{mean} < T_{3},
    \end{cases}
\end{equation}

where $T_{mean}$ denotes daily mean temperature. In other words, we take a "high-end" cut-off.
However, one can also explore and experiment with models involving more complex cut-offs (both, low and high end), as well as heat stress accumulation (i.e. taking the account of consecutive hot days). For further details, see Subsection \ref{subsec:appendix-det-heat}.

\subsubsubsection{\textbf{Deterministic drought stress (the effects of warmer and colder drought)}}

The following simple concept lies at the heart of our deterministic drought model: if the available soil water ($\operatorname{AW}$) is below a certain critical threshold (the so-called maximum allowable depletion $(\operatorname{MAD})$), then the plant starts to experience drought stress. We refer to Subsection \ref{subsec:appendix-det-drought} for a computing $\operatorname{AW}, \operatorname{MAD}$.

Using the quantities (MAD) and (AW), we define our initial deterministic daily drought stress as:

\begin{equation} \label{eq:det-drought-stress}
    S_D := \left[ \left( \operatorname{MAD} - \operatorname{AW} \right) \left( 1 - q \cdot \operatorname{IRR} \right) \right]_+,
\end{equation}
where $q \in (0, 1)$; the variable $\operatorname{IRR}$ assumes integer values in $[0, 3]$ and represents the irrigation treatment (0 corresponds to NULL and 3 corresponds to FULL irrigation); and where $[\cdot]_+$ is the usual positive part (ReLU) function.

A few comments are in place.

First, the irrigation treatment appears at the very last step in the model: depending on our choice of $q$ one could assume that reasonable irrigation is properly conducted so that $S_{drought}$ vanishes (e.g. $q = 1/3$ and full irrigation $\operatorname{IRR} = 3$); or that irrigation might still not entirely suffice to eliminate $S_{drought}$ (e.g. if $q = \varepsilon + 1/3 $, then even full irrigation $\operatorname{IRR} = 3$ might leave $\varepsilon$-traces of drought stress provided there are severe precipitation deficiencies). 

Second, at this point we focus to a large extent on modeling drought for slightly warmer (beyond snow melting point) time periods. In other words, in our initial drought model we do not discuss the effect of snow water equivalents - one should keep in mind that for approx. 81 of our 1561 environments, there is snow presence in the active planting time.

Lastly, we particularly explore variants of the deterministic drought stress for environments which achieve warmer daily temperatures (e.g. above $37 \, C^0) $ - thus, one expects to discern between the effects of (hot, dry) conditions opposed to (cold, dry) conditions. For further details, see Subsection \ref{subsec:appendix-det-drought}.

\subsubsubsection{\textbf{Stresses due to combined heat and drought}}

According to the sources mentioned above, in extreme environmental conditions heat and drought are often coupled - thus, their individual effects might be somewhat challenging to disentangle.

First, in an attempt to disentangle the effects of heat/drought stresses, as already mentioned in the short deterministic heat/drought stress descriptions above, we consider various reactions due to warmer/colder drought and hot/cold stress. For further details, we refer to Section \ref{sec:Appendix}.

However, concerning the combined effect of heat/drought stresses, we initially use the following straight-forward deterministic models:

\begin{equation} \label{eq:combined-stress}
    S_{HD} := f(S_H, S_D),
\end{equation}
where we select different choices for $f$ - products, sums and exponentials thereof. For further details, we refer to Subsection \ref{subsec:appendix-Combined}.

\subsubsubsection{\textbf{From daily stresses to growth-stage stresses}}

So far, the DEM models produced daily stress values - it is reasonable to have a more global aggregated stress values based on growth-stages (cf. Subsection \ref{subsec:appendix-GDUs} for further background and motivation). A straight-forward way to achieve this is the following: the Heat/Drought stresses are summed over growth stages - thus, for each planting instance we have obtained two $18$-dimensional vectors $S_H, S_D$ (where the components correspond to the growth stages).

\subsubsection{Data-driven models}
\label{Data-driven-models}
We use data-driven methods to learn a novel representations of environmental stresses. We model the heat stress and the drought stress with convolutional neural networks, which takes the weather and the performance data as input. A purely 'black-box' approach would, however, be not favorable due to the lack of interpretability. To overcome this obstacle, we combine the neural networks with the expert knowledge through feature selection, feature engineering and informed model building. 

We need to ensure that the two stress modules learn different representations, i.e., one for heat and one for drought stress. To achieve this, we perform a feature selection according to the expert knowledge: The heat module exclusively takes TMAX, TMIN and TMEAN as features while the drought model takes PREC, SWE and VP. Additionally both modules share the following features: CLAY, SILT, SAND, AWC, PH, OM, CEC, KSAT, DAYL, SRAD. We denote the total number of features by $N$.

To further exploit expert knowledge we use the calculated GDU (\ref{eq:daily-GDU}) as an input feature to both models, allowing the neural networks to learn from the growth period of the plant during the season.

The data consists of daily (weather data) as well as static data (environmental data). Let us consider one planting instance of a hybrid. For each day between planting and harvesting we concatenate the weather data with the static data. This gives an input matrix $X_{dj}$, where $d$ enumerates the days for this hybrid and $j\in \{ 1,\dots,N \}$ enumerates the different features. 

 The convolutional neural networks (cf. \cite{NIPS1989_293}) have only one filter mask which moves along the daily dimension $d$. The filter width is equal to the input space dimension $N$ and it has variable height $h$ and variable stride size $s$. 
 The output from the convolution is passed trough a non-linear activation function. The choice of the activation function again is informed by expert knowledge: For the heat module we us an RBF kernel with a trainable variance, reflecting the bell shape of the expert model for heat (Sec. \ref{subsec:appendix-det-heat}). The drought stress is passed trough hyperbolic tangent function. 
 
\begin{figure}
    \centering
    \includegraphics[width=8cm]{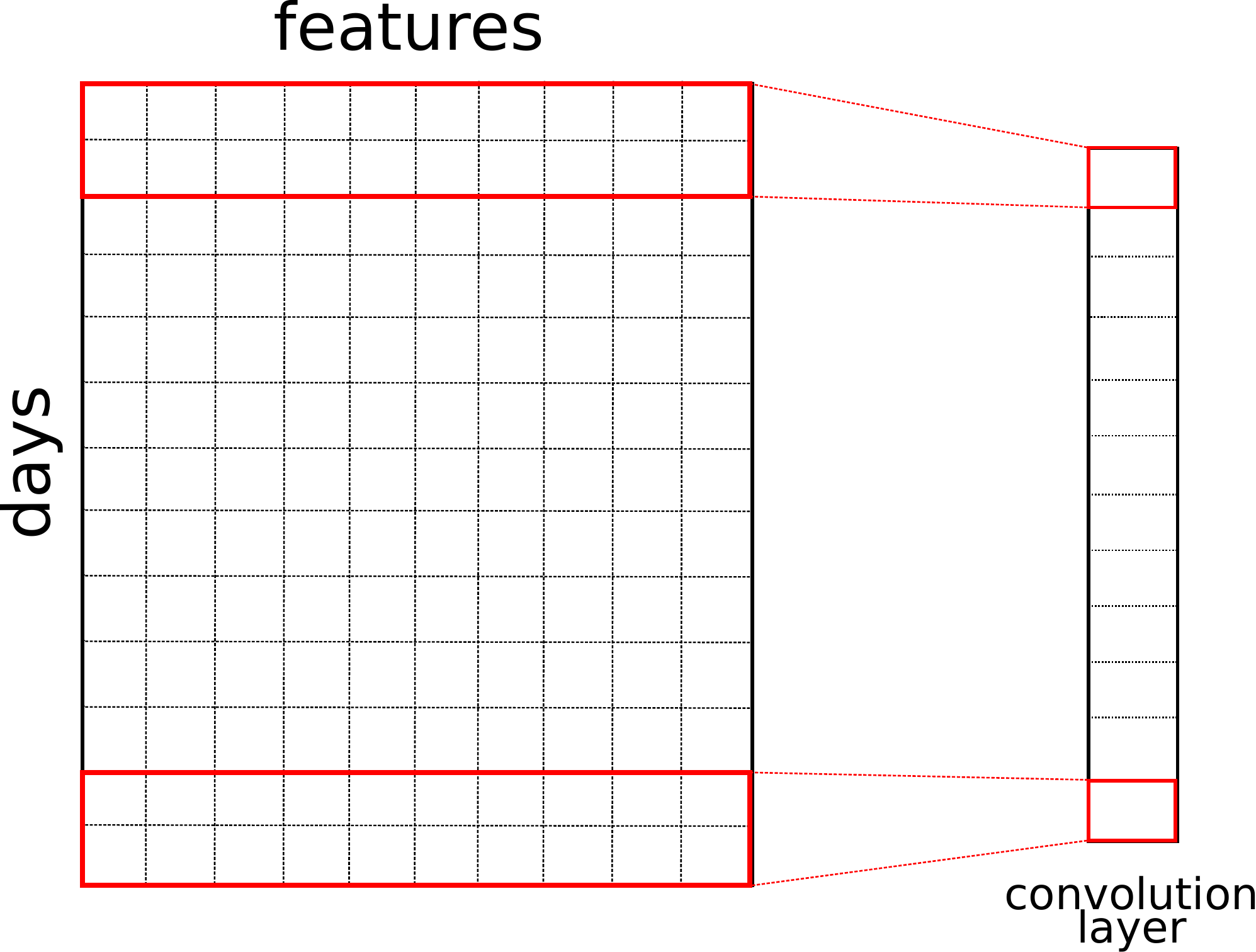}
    \caption{The filter mask moves along planting days - stresses are aggregated quantities
over consecutive days}
    \label{fig:conv_filter}
\end{figure}

 After the convolution and the non-linearity we obtain the stress vector $S_t$, where the range of $t$ depends on the filter size. Since we only use one filter, $t$ can be seen as an averaged time axis, allowing us to interpret the sensitivity of hybrids in a time-resolved manner. This procedure is depicted in Fig. \ref{fig:conv_filter}.
 
 In conclusion, we obtain two stress vectors $S_{\{H,D\}}$. As before, we study the interaction of the two stresses by the combined feature $S_{HD}=f(S_H,S_D)$.

\subsection{Regression model} \label{sec:Regression-Model}

In order to test the designed stress features and their reliability to infer yield losses, we consider a regression on the delta-yield ($\Delta Y$), i.e. the difference between the current observed yield ($Y$) and the maximum yield that the current corn hybrid is able to achieve over all its planting instances. Naturally, high levels of stress should imply higher off-sets from the plant's maximum yield potential, i.e. higher $\Delta Y$.

We train a multi-layer perceptron (MLP) to regress on the $\Delta Y$ from the observed stresses and the corresponding hybrid id $i$ and enviroment id $j$, e.g., we have the model:

\begin{equation}
    \Delta Y = \operatorname{MLP}\left( S_H, S_D, S_{HD}, i, j \right).
\end{equation}

\subsection{Sensitivity measures}

The final goal is to investigate the sensitivity of the hybrids against stresses. To this end we define two sensitivity measures:
\begin{itemize}
    \item Sensitivity based on correlations,
    \item Sensitivity based on derivatives of the MLP with respect to stress features.
\end{itemize}

\subsubsection{Correlation Sensitivity: Co-variance Matrix between Yield and DEM-Stress models}

We define the following covariance matrix $C_{it} \in \mathbb{R}^{2452 \times 18}$, whose dimensions correspond to the number of different corn hybrids (a total of 2452) and the number of periods (a total of 18).

In a nutshell, $C_{it}$ keeps track how the stress $S$ experienced by the corn hybrid $i$ in the period $t$ covariates with the acquired yield. Here covariance is taken w.r.t. all the planting instances of hybrid $i$.

More formally, let us fix a corn hybrid $i$. From the performance data we extract all the rows corresponding to planting instances of hybrid $i$ - say, we have $s$ many rows $x_p^{i}, p = 1,\dots, s$. We note that each planting instance $x_p^{i}$ has a reported delta-yield ($\Delta Y$) quantity denoted by $x_p^{i}[\Delta Y]$ (see Section \ref{sec:Regression-Model} for a definition of $\Delta Y$).

Now, let $S$ be one of our deterministic daily stresses (either heat or drought). For each planting instance $x_p^i$ (row) we have a daily stress vector $S_p$ (computed with respect to the corresponding planting environment). We sum the components of $S_p$ period-wise (i.e. according to growth stages) to obtain a vector $s_p \in \mathbb{R}^{18}$ - each component of $(s_p)_t, t = 1, \dots, 18$ represents the net amount of stress in the period (growth stage) $t$.

Finally, we define our covariance matrix as

\begin{equation}
\label{DEM_sensitivity}
    C_{it} := \operatorname{cov}_{p} \left( x_p^i[\Delta Y], (s_p)_t  \right), \quad \text{for every hybrid id } p \text{ and every time period }t. 
\end{equation}

Here, the covariance is taken with respect to the index $p$, i.e. along all planting instances.

\subsubsection{Derivative Sensitivity: Susceptibility matrix}

For each hybrid $i$ we have a set of planting instances denoted by $p(i)$ as before. For each instance we can evaluate the derivative of the regression function with respect to the stresses by simply back-propagating the activity through the MLP. To get an averaged behaviour for one hybrid we sum this partial derivatives. This gives a sensitivity matrix for the heat stress, the drought stress and the combination thereof:

\begin{equation}
\label{CNN_sensitivity}
    R^{\{D,H,DH\}}_{it} := \sum_{p(i)}\frac{\partial{\Delta Y}}{\partial{S^{\{D,H,HD\}}_t}} 
\end{equation}

\subsection{Ranking and clustering}

In order to form a basic susceptibility ranking of the corn hybrids, we use the sensitivity metrics $C$ or $R$ in the following way. Row-wise we compute a certain vector norm (e.g. $L^1$, $L^2$, $L^\infty$) and rank the rows (i.e. the different hybrids) according to this norm. High levels of the norm should indicate greater susceptibility to stress. In order so solve the final clustering objective we perform a K-means clustering on the rows of the susceptibility matrix $R^{\{D,H,DH\}}$.

\subsection{Experimental Setup}

We train the CNN-MLP as well as the DEM-MLP models in an end-to-end fashion by gradient descent on the mean square error loss function using the ADADELTA. Note, that in the DEM module all parameters are fixed. 

For the CNN model we tested various combinations of filter sizes and strides as well as different number of layers in the MLP. The final model used in the main manuscript (selected according to \ref{model selection}) has the following parameters: The height is $h=15$ and the stride is $s=12$. The MLP has 3 layers with dimensions 64, 80, 40. RELU is used as activation function for all layers. The combination function $f$ is chosen to be the element-wise product of the two stress vectors. 

The parameters of the DEM-MLP model are chosen as follows. For the DEM models, we set the parameters in the heat stress $S_H$ (Eq. \ref{eq:det-heat-stress}) to $T_2 = 25, T_3 = 30$ in accordance with the literature. Furthermore for the drought stress (\ref{eq:det-drought-stress}) we use all environments (both warm and cold). The dimension of the heat and drought vectors is $18$ (Sec. \ref{DEM-models}). The input dimension to the MLP is $3*18 + 2$ (resp. for $S_H, S_D, S_{HD}$ and the hybrid id and environment id). In consequence the 3 layers have dimensions, 56, 30, 20.

\section{Quantitative Results}
In this section we provide the quantities results for both objectives, the regression on the yield reduction as well as the clustering of the hybrids.

\subsection{Regression}
To compare the predictive capabilities of the DEM-MLP and the CNN-MLP model we consider the mean square error (MSE) of the prediction on a test set. To better judge the performance we divide the error by the standard deviation of the data $\sigma = \operatorname{std}(\Delta Y)$ (Tab. \ref{tab:table1}). Both models have a relative error significantly smaller than one, implying that both models are able to predict the yield reduction beyond trivial statistics of the data. The data-driven model has a lower error rate which could be explained as follows: The parameters, e.g. the thresholds, of the DEM stress models might not fit well to the given data while the data-driven is able to learn such thresholds from the data.

\begin{table}[!ht] \label{tab:Performance-Regressor}
  \begin{center}
    \caption{Predictive performance of expert and data-driven models}
    \label{tab:table1}
    \begin{tabular}{l|c|c|r} 
      \textbf{Model} & \textbf{$MSE/\sigma$} \\
      \hline
      DEM-MLP & 74.0 \% \\
      CNN-MLP & 57.1 \% \\
    \end{tabular}
  \end{center}
\end{table}

\subsection{Temporal behavior}
In this subsection we consider the sensitivity of the hybrids to the stresses occurring during the growth period. 

\begin{figure}[!t]
\centering 
\includegraphics[width=14cm]{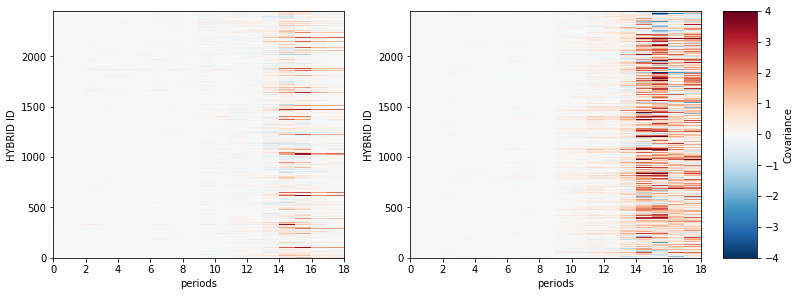}
\includegraphics[width=14cm]{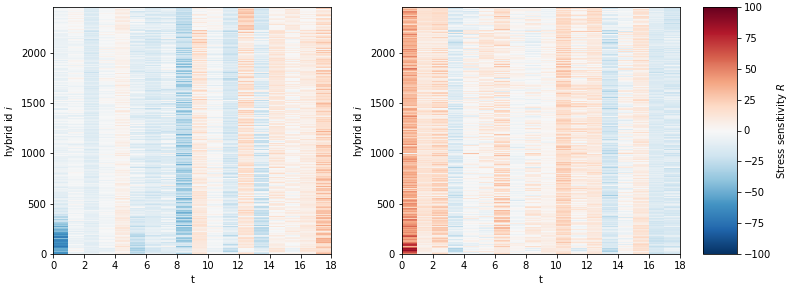}
\caption{Sensitivity analysis. Upper row: Covariance matrices (\ref{DEM_sensitivity}) for the heat $C_H$ (left) and for the drought $C_D$ (right) for the DEM stress model. Lower row: Susceptibility matrices (\ref{CNN_sensitivity}) for the heat $R_H$ (left) and for the drought $R_D$ (right) for the CNN stress model.}\label{fig:temporal}
\end{figure}

For the DEM stress model we use the co-variance matrix $C$ (Eq. \ref{DEM_sensitivity}) between the yield reduction and the stress (Fig. \ref{fig:temporal}). For both stresses we observe that the hybrids suffer from stresses beginning from period 14. This periods, according to the literature and the way we built our deterministic models, correspond to the vital pollination periods (Sec. \ref{subsec:GDUs}). The observed sensitivity in this periods is in line with the agronomy literature - indeed, the time frame around pollination seems to be the most delicate one and sensitive to undesirable environmental conditions.

For the CNN stress model we show the sensitivity matrix $R$ (Eq. \ref{CNN_sensitivity}). Here, the columns reflect time since they are the result of the convolution along the days (\ref{Data-driven-models}). Therefore, a direct comparison to the DEM model has to be taken with reservation, because in the DEM model the periods are based on the accumulated GDU instead of just time. We first consider the heat stress and observe a sensitive time interval near the end of the growing period, which is similar with the DEM model. On the other hand, sensitive time intervals also occur earlier in the growing period, showing that the data-driven approach uncovers relationships which are ignored by the deterministic expert models. For the drought stress the sensitive time intervals appear throughout the growing season except near the end; an opposite behavior compared to the DEM model.

A further investigation of this patterns and in particular the differences is beyond the scope of this report. We nevertheless conclude that both models pick up interesting patterns shedding light on the sensitive regions of the plants during growth.

\subsection{Ranking and clustering}\label{model selection}

In this subsection we consider the sensitivity of individual hybrids and finally cluster the plants into susceptible and not-susceptible hybrids.

\subsubsection{\textbf{Comparison and CNN model selection}}
For the data-driven models we tested various parameter combinations (filter sizes, neuron numbers, number of layers), which all result in similar predictive performance. However, the clustering task is unsupervised and thus there is no direct test whether the learned stress representations are meaningful. To ensure a meaningful representation we check whether the stress sensitivity of individual hybrids obtained by the CNN model resembles to some extent the DEM model.

\begin{figure*}[!ht]
\centering 
\includegraphics[width=7cm]{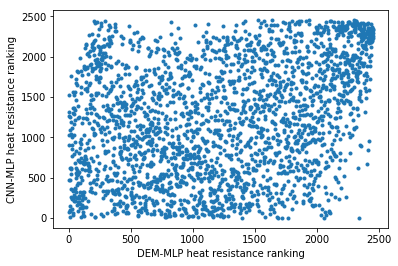}
\includegraphics[width=7cm]{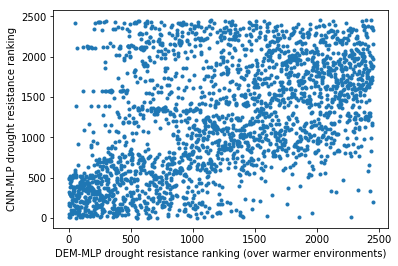}
\caption{Ranking comparisons between the CNN and the DEM model for heat (left) and drought (right) sensitivity. Each point corresponds to one hybrid and the position in one ranking is plotted against the position in the other ranking: points close to the identity line signify similarity of the rankings.}\label{fig:comparision}
\end{figure*}

Let us first consider the heat stress sensitivity. We rank the hybrids according to the sum of their respective rows in the sensitivity matrices $S_H$ (for CNN model) and $C_H$ (for DEM model) and compare the CNN- to the DEM-rankings  (Fig. \ref{fig:comparision}). We have tested different hyper-parameters of the CNN-model (not shown) and pick the model which has the strongest similarity to the DEM model.

A corresponding analysis is conducted for the drought stress. A quite intriguing drought ranking correlation between the DEM and CNN models is obtained when one considers DEM drought stress models for warmer environments (cf. Fig. \ref{fig:comparision}). In this analysis we have also explored various DEM-rankings (induced by changing the thresholds in the DEM models, not shown here). For more details, we refer to Sec. \ref{subsec:appendix-det-drought}.

 Note that this is meant to be only a consistency check: The models should resemble each other only to some extent.

\subsubsection{\textbf{Clustering}}

\begin{figure*}[!ht]
\centering
\includegraphics[width=16cm]{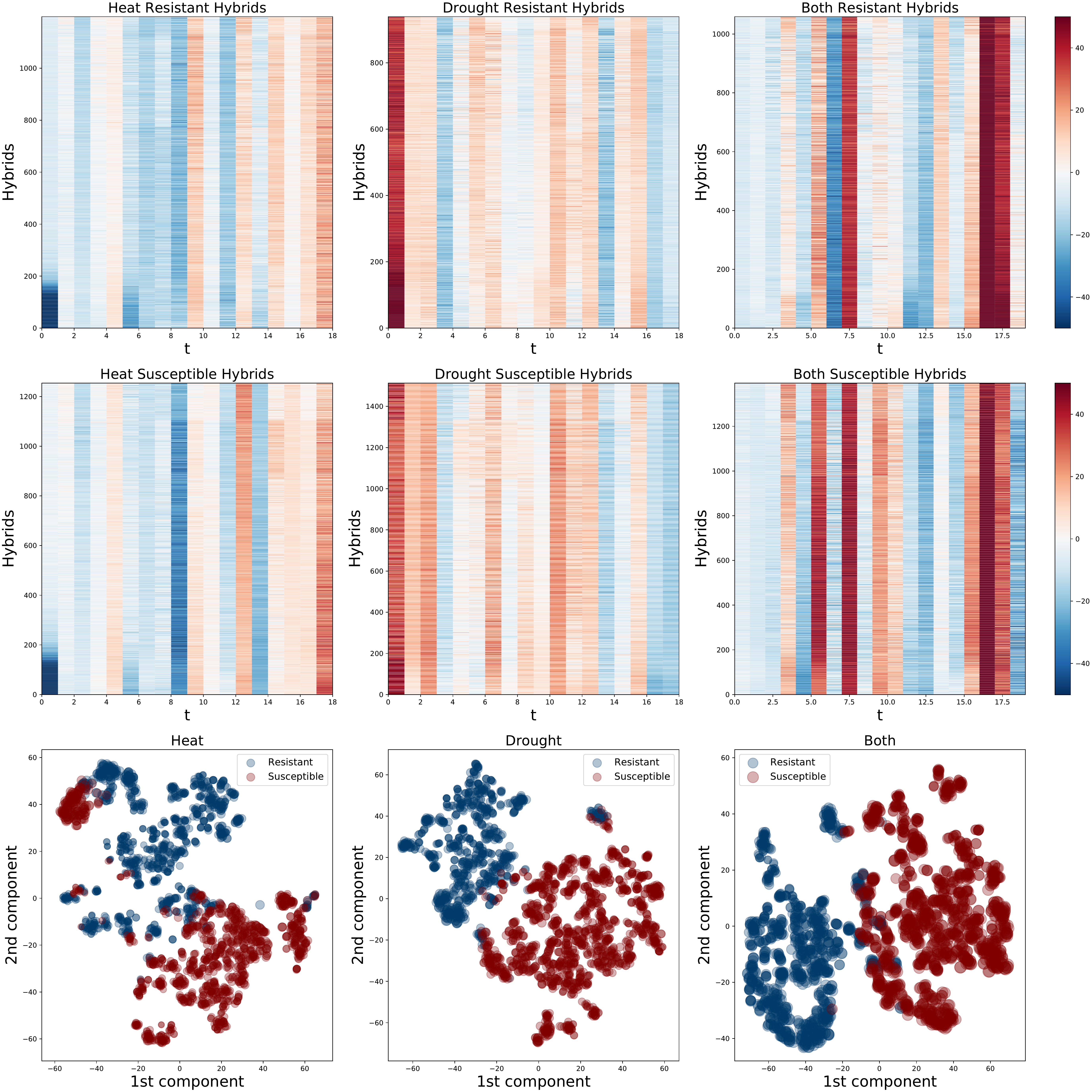}
\caption{Clustering of the hybrids into susceptible and resistant ones. Upper row: Sensitivity matrix only shown for resistant hybrids (their corresponding rows) for heat stress (left), drought stress (middle) and combined stress (right). Middle row: The corresponding values for the susceptible hybrids. Lower Row: A dimensionality reduction (TSNE) of the clustering showing a clear separation between the clusters.}\label{fig:clustering}
\end{figure*}

Finally, we perform a K-means clustering on the rows (the hybrids) of the sensitivity matrix $R$ obtained from the CNN-MLP model (Fig. \ref{fig:clustering}). For each stress we obtain a distinction (silhouette coefficients: heat: 0.25, drought 0.27, both 0.28) into two clusters and identify the one with the higher averaged sensitivity values as the susceptible class. A detailed analysis of the temporal patterns appearing in the two clusters is beyond the scope of this report. To conclude we observe that $36\%$ of the hybrids are resistant to all stresses.

\section{Conclusion}
In this work we develop a data-driven stress representation based on weather and performance data. To this end we use the representation capabilities of convolutional neural networks. To ensure that these representations are meaningful we use expert driven models as an inspiration (e.g. for feature selection) and as a benchmark (for model selection).

Our model is able to predict the yield reduction with high accuracy (Objective 1). Furthermore a sensitivity analysis of the regression function with respect to the stresses allows a clustering of the hybrids into susceptible and resistant ones.

\bibliographystyle{unsrt}

\newpage
\section{Supplementary Materials (Optional)} \label{sec:Appendix}
\subsection{A remark about units, measurements, conversions and constants.}

At certain points in our DEM models, one should keep in mind what the measurement units are.

Although not explicitly given in the data, one can use appropriate considerations to determine what units are used to represent the environmental/performance data - e.g. in order to detect the elevation measurement units, one could consider the geographic coordinates.

In order to avoid explicit technicalities stemming from conversions (e.g. constant interplay between cm. and inches), in the following we do not discuss particular measurement units. Thus, we hope that the exposition becomes clearer and the main ideas/models are more apparent.

However, for our implementations we carried out the necessary conversions.

Finally, in what follows there will be certain physical constants (e.g. specific heat at constant pressure, etc) which will directly be approximated by decimal numbers in the equations (e.g. $0.408$, etc) without further notice. Discussing each appearing physical constant seems to be out of the scope of this text. For details, we refer to \cite{JBA}.

\subsection{A few remarks about environments, time periods, GDUs and growth stages.} \label{subsec:appendix-GDUs}

\textbf{A brief motivation}. On one hand, it is convenient to discover the different stages of our planting instances as this will uniformize the time frames - instead of constantly working with the varying number of crop days for each corn planting instance, we will have a globally uniform time scale based on the maturity stages.

On the other hand, detecting maturity periods (especially pollination) will better reflect the impact of heat/drought stresses. Many agricultural works (cf. \cite{AEB}, \cite{RWG}, \cite{L}, \cite{ET}, \cite{WHN}) explicitly claim that certain corn growth stages are far more delicate and sensitive to stress than others - further, the emphasis is heavily placed upon the stages of pollination and grain-fill, which lead to significant yield losses when damaged by severe heat/drought conditions. 

\textbf{Computing GDU, accumulated GDU and growth stages}. First, we briefly describe the following property. One can assume that a lot of performance-related features are actually environment-related features. More precisely, note for each two rows $x_i, x_j$ of the performance data (i.e. $x_i, x_j$ represent different planting instances) such that $x_i[\operatorname{ENV\_ID}] = x_j[\operatorname{ENV\_ID}]$, there holds
\begin{equation}
    x_i[\operatorname{PLANTING\_DAY}] = x_j[\operatorname{PLANTING\_DAY}]
\end{equation}
\begin{equation}
    x_i[\operatorname{HARVEST\_DAY}] = x_j[\operatorname{HARVEST\_DAY}].
\end{equation}

In other words, each environment has a globally fixed planting/harvest days which are invariant of the planting instance. The same holds for the all the performance features (soil properties, etc), except the features related to yield.

We can thus reduce the environmental data by only considering the plant/harvest days for each environment.

For each environment we compute a daily GDU by means of eq. (\ref{eq:daily-GDU}), after which we compute a normalized daily accumulated GDU (AGDU) given by:

\begin{equation}
    \operatorname{AGDU}_d :=\frac{1}{ \operatorname{AGDU}_{\text{Harvest Day}}} \left( \sum_{i = \text{Planting Day}}^d \operatorname{GDU}_i \right),
\end{equation}
where lower indices denote the corresponding day number. Note that $\operatorname{AGDU}_d$ is a number in the interval $[0, 1]$.

Finally, inspired by the available literature (see, for example, \cite{AEB} and the references therein), we can subdivide the interval $[0, 1]$ into the following sub-intervals corresponding to different growth stages and sub-stages:

\begin{align*}
    \{P_i\}_{i=0}^{17} =& \{ (0, 0.023111), (0.023111, 0.04411), (0.04411, 0.085111),\\ &(0.085111, 0.1075111), (0.1075111, 0.1231111), (0.1231111, 0.1334111), \\
    &(0.1334111, 0.1531111), (0.1531111, 0.1635111), (0.1635111, 0.1660111), \\ &(0.1660111, 0.1925111), (0.1925111, 0.22111), (0.22111, 0.27111), \\ &(0.27111, 0.3111), (0.3111, 0.4111), (0.4111, 0.6111), \\ &(0.6111, 0.8111), (0.8111, 0.9111), (0.9111, 1) \}.
\end{align*}

Here we remark that we work in a framework with $18$ time periods - a further refinement of the agronomically described time frames. One expects that the above periods capture crucial maturity stages and sub-stages such as emergence, pollination, tasseling, silking, etc.

\subsection{Drawbacks of DEM models} \label{subsec:appendix-DEM-drawbacks}

When dealing with the mentioned deterministic expert-driven models, we were faced with the following challenges:

\begin{enumerate}
    \item \textit{"Manually fine-tuning a zoo of parameters"}. Considering the thorough survey in \cite{Zhe1}, it seems that the more complicated and involved the stress models are (thus, perhaps, the more accurate), the more parameters they actually contain: these include various further environmental data (e.g. canopy information), physical threshold parameters and regime-switches, crop(corn)-specific parameters, various types of cut-off and bump functions, etc.
    
    Thus, the question arises: How should one select the best of these models? Further, how should one select the optimal parameters having the data at hand? A first solution might be to go for a couple of specific models and optimize the parameters via an appropriate ML method (e.g. by regressing yield quantities with respect to the stress models). However, this raises further issues in terms of model interpretability.
    
    \item \textit{Uncovering hidden knowledge and model complexity}. The more underlying connections we would like to utilize or discover, the more non-trivial our model becomes - further, this has to be executed with the data at hand. This usually leads to highly non-linear simulations with high complexity. In other words, one faces the trade-off between tractability and accuracy.
\end{enumerate}

\subsection{Further details: deterministic heat stresses} \label{subsec:appendix-det-heat}

Here we provide further information on the design of our deterministic heat stresses.

\textbf{Temperature stress.} First, let us recall that the APSIM Maize model (cf. [\cite{Zhe1}]) aims to capture the daily biomass production as function of several quantities among which, a certain daily temperature stress, $S_{temp}$, computed as the following linear cut-off function: 

\begin{equation}
    S_{temp} := 1 - \begin{cases} 0, \quad \text{if } T_{mean} \leq T_0 \text{ or } T_{mean} \geq T_3,  \\
                \frac{T_{mean} - T_0}{ T_1 - T_0}, \quad \text{if } T_0 < T_{mean} < T_1, \\
    1, \quad \text{if } T_1 \leq T_{mean} \leq T_2, \\
    \frac{T_3 - T_{mean}}{T_3 - T_2}, \quad \text{if } T_2 < T_{mean} < T_{3}.
    \end{cases}
\end{equation}

Here, $T_{mean}$ denotes the daily mean temperature, and $T_i, i = 0, 1, 2, 3,$ denote different temperature thresholds.

We note that the cut-off function $S_{temp}$ captures both the effect of extremal hot and cold temperatures.  

Further, it seems that tuning the $T_i$'s could be a delicate matter - in the "default" APSIM model, the thresholds are set to (8, 15, 35, 50); however, in [\cite{Zhe1}], these are assumed to be (8, 15, 30, 44) as the model would thus be more responsive to lower temperatures. In our case, the data we are dealing with involves a large proportion of relatively colder environments so it is reasonable to adjust the thresholds even further. We also refer to Subsection \ref{subsec:appendix-DEM-drawbacks}.

\begin{figure*}[!ht]
\centering 
\includegraphics[width=10cm]{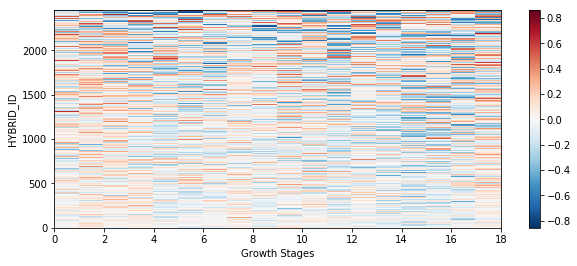}
\caption{DEM $S_{temp}$ stress model: sensitivity analysis via correlation. The rows are already sorted according to their Euclidean $L^2$-norm thus giving a resistance ranking of the hybrids. The effects of colder temperatures is also captured.} 
\label{fid:temp-sensitivity}
\end{figure*}

We have explored several different threshold parameters for the cut-off function - among them the choice
\begin{equation}
    \{T_i^1\}_{i=0}^3 := \{ 10, 20, 25, 35 \},
\end{equation}
appears to be well-responsive with the data at hand, giving a well-behaved and more uniform yield covariance matrix $C_{ip}$ defined by equation (\ref{DEM_sensitivity})  - we refer to Fig. \ref{fid:temp-sensitivity}.

\begin{figure*}[!ht]
\centering 
\includegraphics[width=7cm]{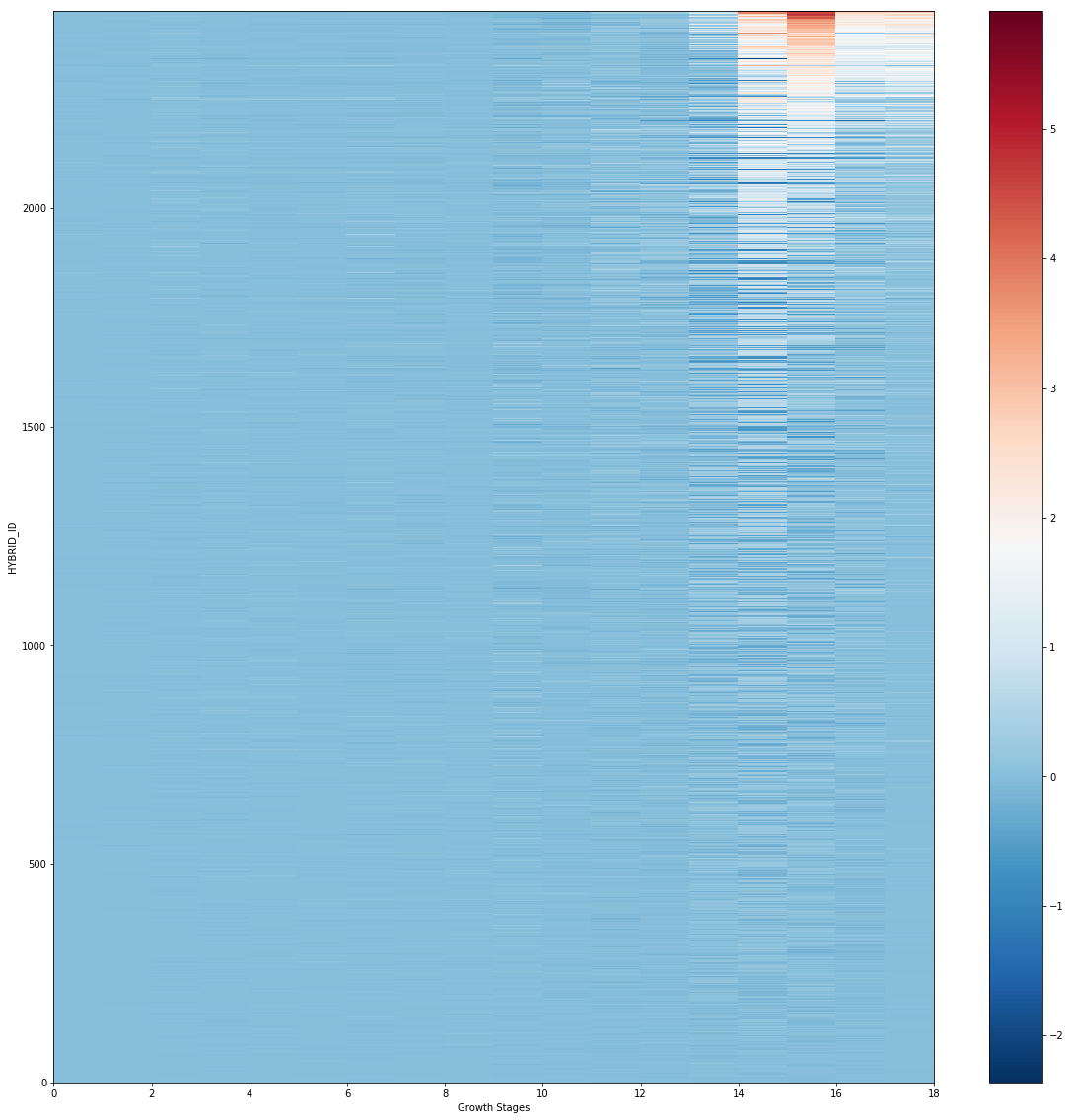}
\includegraphics[width=7cm]{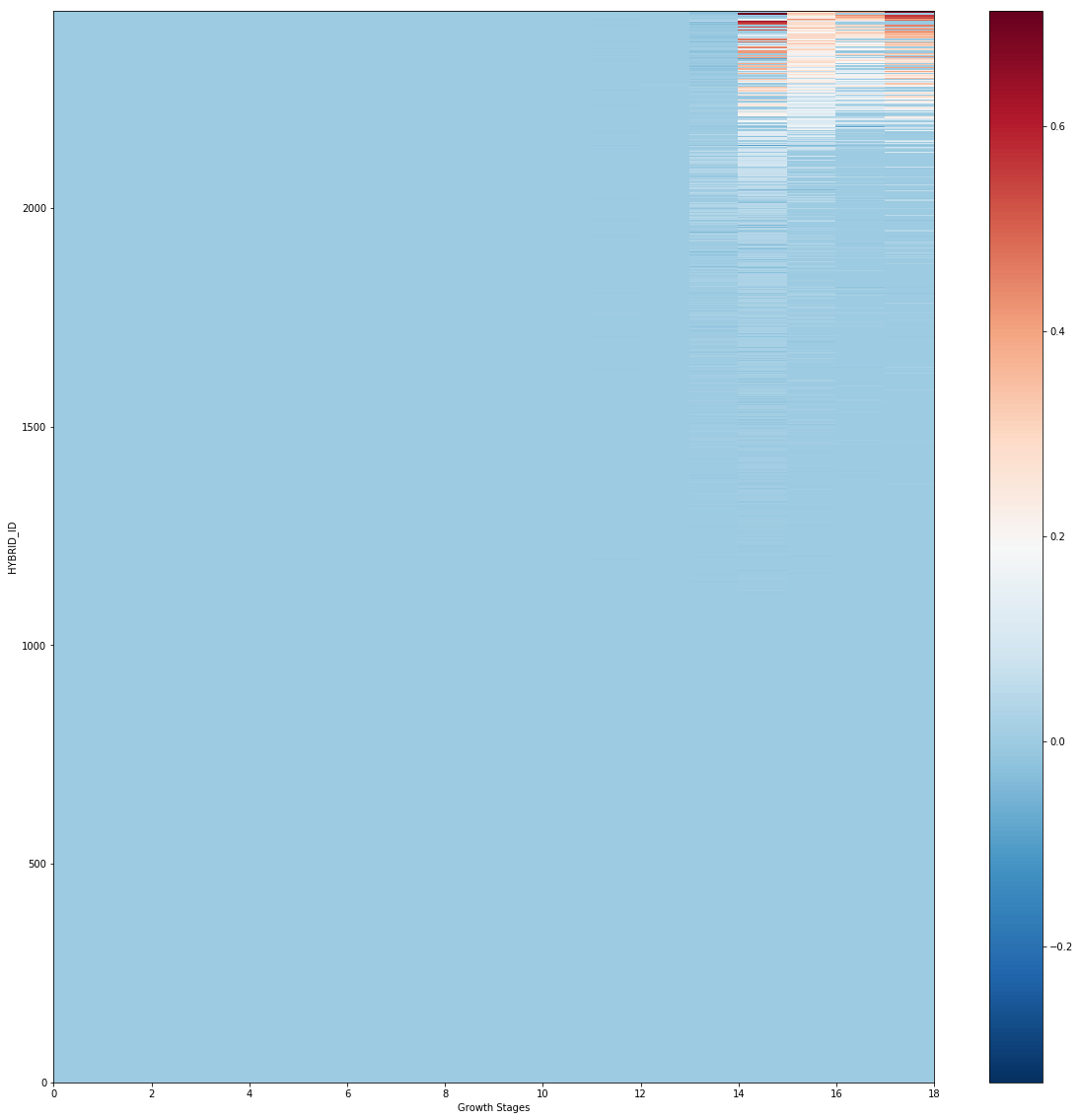}
\caption{DEM $S_{H}$ stress model: sensitivity analysis via correlation. Left: thresholds $T^1_i$; Right: thresholds $T^2_i$. The rows are already sorted according to their Euclidean $L^2$-norm thus giving a resistance ranking of the hybrids. The effects of colder temperatures are not captured that prominently.} 
\label{fig:heat-sensitivity}
\end{figure*}

\textbf{Heat stresses (high-end cut-off).} We recall the definition of $S_{H}$ from equation (\ref{eq:det-heat-stress}). According to several sources (cf., for instance, \cite{ET}, \cite{RWG}) it is reasonable to view the threshold of $30 \, C^0$ as a point beyond which heat stress emerges. Among the various parameter options we have explored, the choices

\begin{equation}
    \{T_i^1\}_{i=0}^1 := \{25, 30 \} \quad  \{T_i^2\}_{i=0}^1 := \{ 30, 35 \},
\end{equation}

seem to reveal a lot of interesting information - see Fig. \ref{fig:heat-sensitivity}.

\begin{figure*}[!ht]
\centering 
\includegraphics[width=10cm]{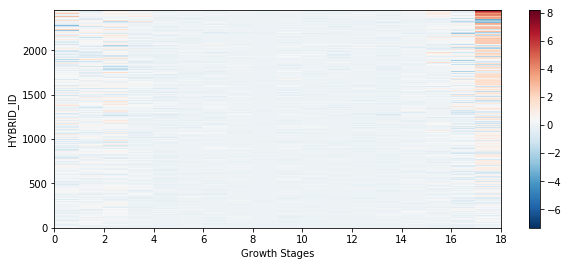}
\caption{DEM $S_{cold}$ stress model: sensitivity analysis via correlation. The rows are already sorted according to their Euclidean $L^2$-norm thus giving a resistance ranking of the hybrids. The effects of hotter temperatures is not so prominently captured.} 
\label{fid:cold-sensitivity}
\end{figure*}

\textbf{Cold stress.} Similarly, we propose the following daily cold stress quantity:

\begin{equation}
    S_{cold} := 1 - \begin{cases} 0, \quad \text{if } T_{mean} \leq T_0,  \\
                \frac{T_{mean} - T_0}{ T_1 - T_0}, \quad \text{if } T_0 < T_{mean} < T_1, \\
    1, \quad \text{if } T_1 \leq T_{mean}, \\
    \end{cases}
\end{equation}
where again we use the same variable notation. To juxtapose the cold stress in contrast with the heat stress, we have used the choice of threshold parameters $(10, 15)$ - see Fig. \ref{fid:cold-sensitivity} for an impression of the sensitivity matrix and corresponding ranking.

\textbf{Accumulation heat stresses.} In accordance with the insights from the present literature, it is reasonable to take into account the effect of the longer uninterrupted exposure to higher temperatures.

Roughly, if the max/min temperatures are high over a longer period of consecutive days, then the heat stress appears to be higher - compared to similar hot periods but with interluded cool-down.

\begin{figure*}[!ht]
\centering 
\includegraphics[width=10cm]{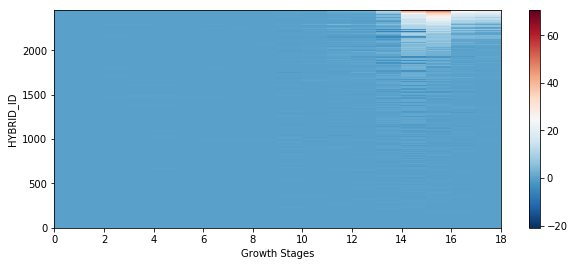}
\caption{DEM $S_{AH}$ stress model: sensitivity analysis via correlation. The rows are already sorted according to their Euclidean $L^2$-norm thus giving a resistance ranking of the hybrids. The effects of hotter temperatures is not so prominently captured.} 
\label{fid:acc-sensitivity}
\end{figure*}

We model this daily accumulated heat stress $S_{AH}$ through:

\begin{equation}
    S_{AH} := A \cdot S_H,
\end{equation}
where $S_H$ is computed via equation (\ref{eq:det-heat-stress}) and $A$ represents the daily accumulation factor, computed in the following way: if $S_H$ has been positive over a consecutive series of $l$-many days, say $d \in \{ d_{0}, \dots, d_l \}$, then $A$ is set to be $\left( 0.9 + (0.1)*l \right) $ for each of these $l$ days. Of course, here the model seems to be rough as the first day of the group receives a similar accumulation factor as the last - this could be easily addressed by a further transformation of $A$. Nevertheless, the model still seems to capture the effect of longer heat exposure. Further, the parameters $0.9, 0.1$ used in the computation can be adjusted. In Fig. \ref{fid:acc-sensitivity} one sees an impression of the accumulated heat stress ($S_H$ is used with $(25, 30)$ as parameters).

\begin{itemize}
    \item \textbf{Comparisons between some rankings of the various DEM heat models}
    We take the time to display how the rankings obtained from the above correlation matrices relate with one another. We refer to Figure \ref{fig:DEM-heat-rankings}.

\begin{figure*}[t!] 
\centering 
\includegraphics[width=8cm]{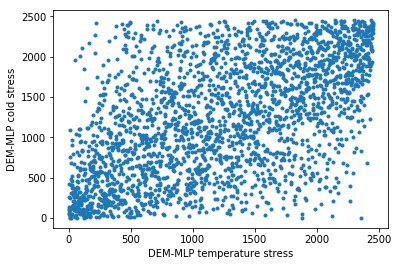}
\includegraphics[width=8cm]{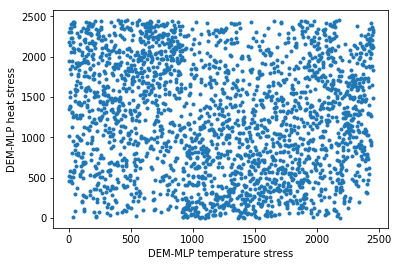}
\includegraphics[width=8cm]{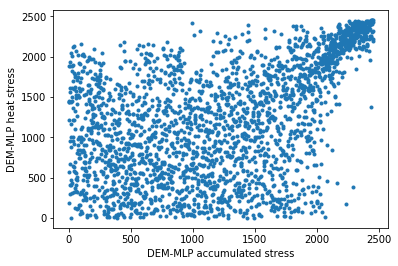}
\caption{Ranking comparison between various rankings obtained by the covariance sensitivity matrix and DEM heat stress models: scatter plots of the pairs of rankings. It is visible that some of the rankings possess partly similar order, whether others do not.}\label{fig:DEM-heat-rankings}
\end{figure*}

\end{itemize}

\subsection{Further details: deterministic drought stresses} \label{subsec:appendix-det-drought}

\begin{itemize}
    \item \textbf{An overview: the maximum allowable depletion threshold (MAD) and the available soil water (AW)}.
\end{itemize}

A central object in our deterministic drought stress analysis is the maximum allowable depletion (MAD) - a threshold describing the soil water level at which drought stress will emerge. Inspired by several ideas from \cite{AC,RY, KID}, we compute the MAD as a certain proportion, $p$, of the plant available water holding capacity, i.e.:

\begin{equation}
    \operatorname{MAD} := p \cdot \operatorname{AWC},
\end{equation}
where $ p \in (0, 1)$, and $\operatorname{AWC}$ denotes the available water capacity, being a function of the soil texture and the corn roots' depth. For details, see the discussion below.

Further, for our computations we need to evaluate the daily available soil water (AW). To this end, we need to take into account effects such as evapotranspiration (i.e. the combined effect of pure soil water evaporation and plant transpiration), soil water runoff and percolation (i.e. downward movement of water through soil) - a thorough overview can be found in \cite{G}. Using these we are able to produce a model of the form:

\begin{equation}
    \operatorname{AW} := \operatorname{AW}\left( \operatorname{PREC}, \operatorname{ET}, \operatorname{RO} \right),    
\end{equation}

where $\operatorname{AW}, \operatorname{PREC}, \operatorname{ET}, \operatorname{RO}$ respectively denote the available soil water, precipitation, evapotransporation and runoff/percolation. Moreover, concerning evapotranspiration, we provide the following (Penman-Monteith-inspired) model:

\begin{equation}
    \operatorname{ET} := \operatorname{ET}(T_{mean}, \operatorname{ELEV}, \operatorname{VP}, \operatorname{SRAD}),
\end{equation}

where $\operatorname{ET}, T_{mean}, \operatorname{ELEV}, \operatorname{VP}, \operatorname{SRAD}$ respectively represent evapotranspiration, daily mean temperature, elevation, vapor pressure and solar radiation.

Lastly, the runoff (RO) is mainly considered as a function of the soil texture and dynamics, i.e.

\begin{equation}
    \operatorname{RO} := \operatorname{RO}( \operatorname{SOIL}).
\end{equation}
Here $\operatorname{SOIL}$ may include quantities such as hydraulic soil conductivity, soil textures, etc.

For a detailed and precise discussion of the models we defined above - we refer to the discussion that follows below.

\begin{itemize}
    \item \textbf{Evapotranspiration in the spirit of Penman-Monteith.}
\end{itemize}
    Having a realistic description of water loss through soil water evaporation and plant transpiration is crucial for our deterministic drought model. To achieve this, we rely on a model similar to a Penman type equation.
    
    We recall that the Penman-Monteith equation is a central and robust tool among the methods of The United Nations Food and Agriculture Organization (FAO) for modelling evapotranspiration (cf. \cite{JBA}). In a nutshell the Penman-Monteith equation should output a so-called reference  evapotranspiration ($\operatorname{ET_0}$), i.e. the rate at which readily available soil water is vaporized from specified vegetated surfaces. We remark that $\operatorname{ET_0}$ depends only on climatic data. However, $\operatorname{ET_0}$ is a reference benchmark and does not necessarily describe correctly the corn evapotranspiration ($\operatorname{ET}$) we actually would like to obtain. To overcome this issue, we normalize $\operatorname{ET_0}$ by a crop specific factor, taking into account the corn growth stages and rough corn water need.
    To summarize we have the following two steps:
    
    \begin{enumerate}
        \item Aggregate the climatic data in an appropriate way and compute reference evapotranspiration $\operatorname{ET_0}$.
        \item Suitably normalize $\operatorname{ET_0}$ to account for corn crops and growth stages.
    \end{enumerate}
    
    Here are the details.
    
    \textbf{Step 1.}

    Our daily reference evapotranspiration is a variant of the Penman-Monteith equation (based on energy balance and aerodynamics considerations) of the form:
    
    \begin{equation}
        \operatorname{ET_0} := \frac{(0.408) \Delta R + \gamma V \frac{900}{273 + T_{mean}}}{\Delta + \gamma}.
    \end{equation}
    
    In the following we elaborate on the appearing terms.
    
    First, $\gamma$ is a psychrometric constant with
    \begin{equation}
        \gamma := (0.000665)P,
    \end{equation}
    where $P$ denotes the atmospheric pressure, which we compute via the elevation $\operatorname{ELEV}$ as:
    \begin{equation}
        P = 101.3\left( 1 - (0.0065)\frac{\operatorname{ELEV}}{293} \right)^{(5.26)}.
    \end{equation}
    
    Further, the quantity $V$ represents the vapor pressure deficit given by
    
    \begin{equation}
        V = (0.61)\exp\left(\frac{(17.27)T_{mean}}{(T_{mean} + 237.3} \right) - \operatorname{VP},
    \end{equation}
    
    with $\operatorname{VP}$ denoting the present vapor pressure.
    
    The term $\Delta$ denotes the slope of saturation vapor pressure and is defined as
    
    \begin{equation}
        \Delta := \frac{(2499.78)\exp\left( (17.27) \frac{T_{mean}}{T_{mean} + 237.3}  \right)}{\left(T_{mean} + 273.3\right)^2}.
    \end{equation}
    
    Finally, the last appearing term $R$ denotes solar radiation.

    This concludes the computation of our daily reference evapotranspiration $\operatorname{ET}_0$

    \textbf{Step 2.}
    
    To compute the actual corn-specific daily evapotranspiration we define the following crop-specific normalization factors (inspired by the mentioned literature):
    
    \begin{align*}
        W := \{W_i\}_{i=0}^{17} =& \{ 1, 1, 2, 2, 2, 3, 3, 4.6, 5, \\ &6, 6, 7, 8, 8, 9.8, 9, 6, 3.5 \}.
    \end{align*}
    
    Thus, we are able to compute the daily evapotranspiration $\operatorname{ET}_d$ for day $d$ as:
    
    \begin{equation}
        \operatorname{ET}_d := W_i \cdot (\operatorname{ET_0})_d,
    \end{equation}
    where the factor $W_i$ is selected depending on which time period (growth period) the current day $d$ belongs to.

\begin{figure*}[t!]
    \centering 
    \includegraphics[width=7cm]{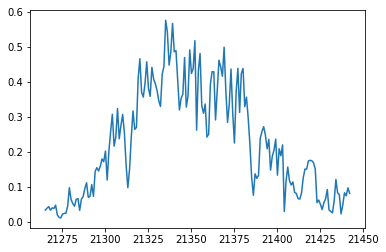}
    \includegraphics[width=8cm]{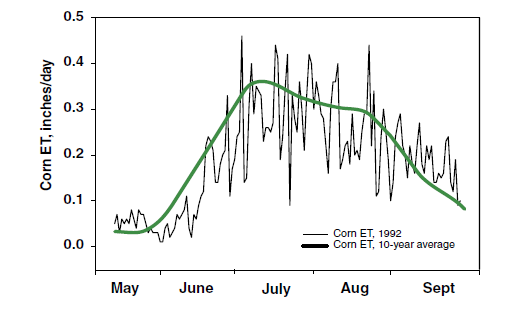}
    \caption{An impression on evapotranspiration (ET): Left: an overview of our ET model for Env\_ID 130; Right: an excerpt of measured ET in Burlington CO, cf. Fig. 26, \cite{BW}.} 
    \label{fig:ET-examples}
\end{figure*}

    For an impression of the present $\operatorname{ET}$ model, we refer to Fig. \ref{fig:ET-examples}.
    
\begin{itemize}
    \item \textbf{Soil dynamics and maximum allowable depletion (MAD).}
\end{itemize}

When one wishes to compute the available water, various further soil characteristics should be taken into account - soil texture determines characteristics such as the capacity to hold water; conductivity; run-off and percolation quantities, etc. For an overview, we refer to \cite{S, BW} and the references therein.

\begin{figure*}[t!] \label{fig:soil-cov}
    \centering 
    \includegraphics[width=10cm]{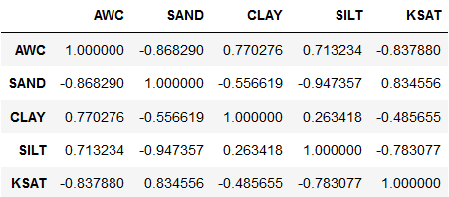}
    \caption{Co-variance between some soil texture parameters.}
\end{figure*}

In the present data we have the available water capacity ($\operatorname{AWC}$) and hydraulic conductivity ($\operatorname{KSAT}$). An immediate co-variance/correlation check (cf. Fig. \ref{fig:soil-cov}) with other available textures parameters such as $\operatorname{SILT}, \operatorname{SAND}, \operatorname{CLAY} $ supports well-known statements from the mentioned literature:
\begin{enumerate}
    \item Sand increases conductivity and reduces the capacity.
    \item Clay and Silt decrease conductivity and increase capacity. 
\end{enumerate}

Thus, to model maximum allowable depletion ($\operatorname{MAD}$) we use the data in a straightforward way and set

\begin{equation}
    \operatorname{MAD} := p \cdot \operatorname{RD}_i \cdot \operatorname{AWC},
\end{equation}

where we select the proportion $p \in (0, 1)$ as $1/2$; $\operatorname{AWC}$ denotes the available water capacity in the present planting instance; $\operatorname{RD}$ is a literature-motivated growth stage dependent parameter describing the corn's root depth:

\begin{equation}
    \operatorname{RD} := \{ \operatorname{RD}_i \}_{i=0}^{17} = \{ 1, 1, 2, 2, 3, 5, 6, 7, 8, 10, 12, 14, 16, 18, 23, 24, 24, 24 \}.
\end{equation}

As in the case of $\operatorname{ET}$, the factor $\operatorname{RD}_i$ in the computation of $\operatorname{MAD}$ is applied according to growth-stage.

\begin{itemize}
    \item \textbf{Computing available soil water.}
\end{itemize}

Finally, we describe how we compute the available soil water, based on precipitation levels ($\operatorname{PREC}$), runoff/percolation ($\operatorname{RO}$) and evapotranspiration ($\operatorname{ET}$).

We address $\operatorname{RO}$ in a straightforward simplistic manner by just considering the hydraulic conductivity $\operatorname{KSAT}$ at hand. We define the daily $\operatorname{RO}$ proportion as:
\begin{equation}
    \operatorname{RO} := f(\operatorname{KSAT}),
\end{equation}
where $f$ is a certain normalization procedure - normalizing $\operatorname{KSAT}$ between $0, 1$ by considering extremal values.

Now, we define the available soil water for day $d$  ($\operatorname{AW}_d$) recursively by:

\begin{equation}
    \operatorname{AW}_d := \left[ \left(1 - \operatorname{RO}\right) \operatorname{AW}_{d-1} + \operatorname{PREC}_{d-1} - \operatorname{ET}_d \right]_+,
\end{equation}
where as usual the function $[\cdot]_+$ is the positive part (ReLU).
The first summand aims to represent the effect of percolation/runoff, whereas the second and the third represent daily precipitation and evapotranspiration.

\begin{itemize}
    \item \textbf{Interpreting the irrigation values.}
\end{itemize}

\begin{enumerate}
    \item \textit{Normal Irrigation}: Assuming that irrigation has been appropriately scheduled (taking in mind, e.g. current crop development and water availability, cf. \cite{WHN,RY}), normally irrigated crops should experience a negligible amounts of drought stress when exposed to no rain or mild/medium rain conditions.
    \item \textit{Light Irrigation and Eco Light Irrigation}: These crops should experience mild/medium drought stress when exposed to long periods of no rain. Increased amounts of precipitation should quickly the levels of drought stress down.
    \item \textit{No Irrigation}: No irrigation implies high levels of drought stress when long-lasting insufficient precipitation is at hand. Here the notion of "long periods" depends on the soil.
\end{enumerate}

\begin{itemize}
    \item \textbf{Drought stress over warmer and colder environments.}
\end{itemize}

A way to alleviate the influence of heat when computing drought stress, could be the following.

Having the above drought stress $S_D$ (cf. equation (\ref{eq:det-drought-stress}) one has a corresponding covariance sensitivity matrix $C$ given by equation (\ref{DEM_sensitivity}).

However, when computing $C$, instead of considering all the planting instances of a hybrid $i$, one could consider only the instances which were planted in a filtered set of environments - e.g. environments which are colder or warmer (i.e. filtered on mean/max/min temperature). Of course, some hybrids might not be planted in a certain set of filtered environments - for these, we set the corresponding row in $C$ to $0$.

In this way, we can discern drought effects in colder/warmer environments. For instance, the warmer environments we consider in most experimental scenarios, are those for which certain days have exceeded mean temperatures of $35, 35$ degrees.

\begin{figure*}[ht!] 
\centering 
\includegraphics[width=8cm]{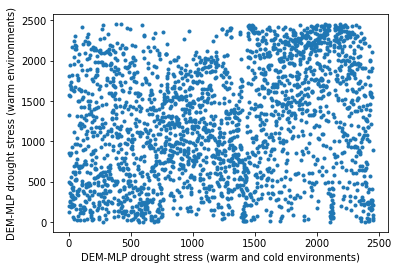}
\caption{Ranking comparison between rankings for different drought stresses (warm versus warm and cold): scatter plot between the two rankings.}\label{fig:DEM-drought-rankings}
\end{figure*}

For an impression on what one can expect in terms of rankings via the covariance sensitivity matrix - see Figure \ref{fig:DEM-drought-rankings}.

\subsection{Further details: combined heat/drought stresses} \label{subsec:appendix-Combined}

Our deterministic combined stress models are defined via the function $f$ (recall (\ref{eq:combined-stress})). Some of the straigh-forward options we explored were:

\begin{equation}
    f(x, y) := \begin{cases} x \cdot y, \\
    x + y, \\
    \exp(x \cdot y) - 1, \\
    \exp(x + y) - 1.
    \end{cases}
\end{equation}

It should be noted that although it captures a combined effect, the sum $S_H$ of $S_D$ should be interpreted as stress in general - it is activated upon the emergence of one stress and increased further if both are present. The product $S_H \cdot S_D$ on the other hand seems to isolate only the combined stress. We also note that taking such a product might result in a rather sparse matrix - so some appropriate renormalization may be needed.

\begin{figure*}[ht!] 
\centering 
\includegraphics[width=9cm]{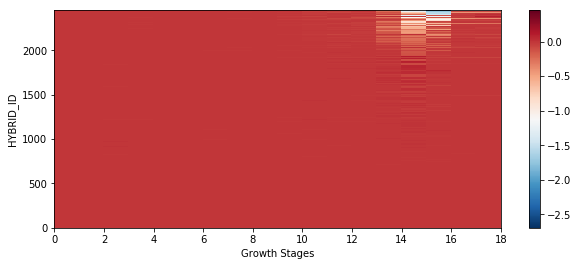}
\caption{Covariance sensitivity matrix: Combined DEM stress with a function f being a product composed with exponential. The rows are sorted with respect to their Euclidean norm.}\label{fig:DEM-combined-sensitivity}
\end{figure*}

For an impression on ranking from the covariance sensitivity matrix we refer to Figure \ref{fig:DEM-combined-sensitivity}.

\begin{figure*}[ht!] 
\centering 
\includegraphics[width=8cm]{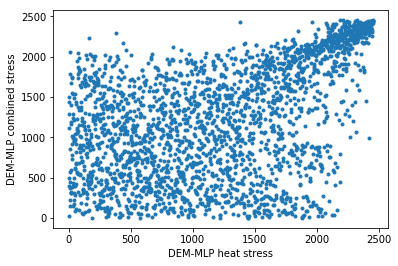}
\includegraphics[width=8cm]{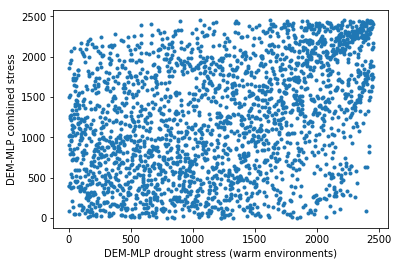}
\caption{Ranking comparison between combined stress and heat/drought stresses: scatter plot between pairs of rankings.}\label{fig:DEM-comb-rankings}
\end{figure*}

Finally, in Figure \ref{fig:DEM-comb-rankings} we see an impression on what the comparison between the rankings of heat/drought and combined DEM models could look like.

\subsection{Additional resistance ranking comparisons}

Lastly, we provide some intriguing further ranking comparisons.

\begin{figure*}[h!] \label{fig:cov-matrix}
\centering 
\includegraphics[width=7cm]{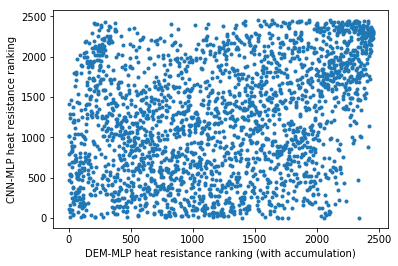}
\includegraphics[width=7cm]{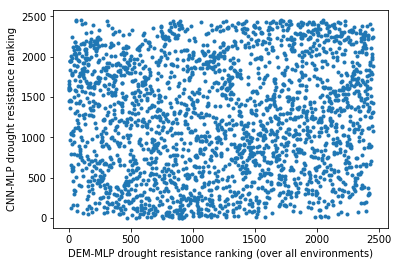}
\includegraphics[width=7cm]{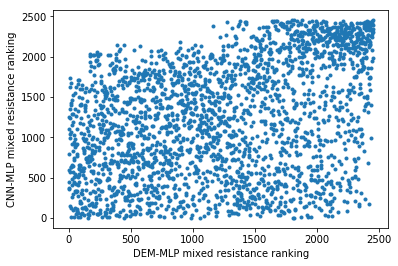}
\caption{Ranking comparisons: Drought resistance vs Heat resistance: scatter plots between different pairs of rankings}\label{fig:further-CNN-DEM-comparisons}
\end{figure*}

In Figure \ref{fig:further-CNN-DEM-comparisons} one can observe further comparisons between resistance rankings obtained by CNN and DEM techniques.

\begin{figure*}[h!] \label{fig:cov-matrix1}
\centering 
\includegraphics[width=7cm]{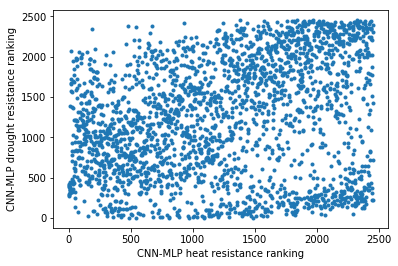}
\includegraphics[width=7cm]{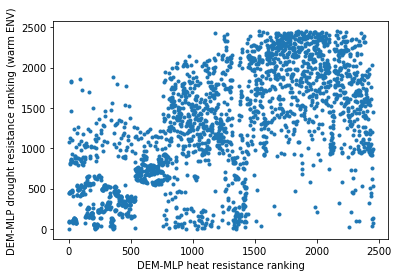}
\caption{Ranking comparisons: Drought resistance vs Heat resistance: scatter plots between different pairs of rankings}\label{fig:further-inter-comparisons}
\end{figure*}

In Figure \ref{fig:further-inter-comparisons} we see how the heat/drought resistance rankings produced by the CNN, resp. by the DEM, techniques relate.

\end{document}